\documentclass[a4paper,11pt]{article}
\usepackage[utf8]{inputenc} 
\usepackage[T1]{fontenc}    
\usepackage{lmodern}        
\usepackage{hyperref}       
\usepackage{url}            
\usepackage{booktabs}       
\usepackage{amsfonts}       
\usepackage{nicefrac}       
\usepackage{microtype}      
\usepackage{graphicx}

\usepackage{algorithm}
\usepackage{algpseudocode}

\usepackage[authoryear]{natbib}
\usepackage[right=2.5cm,left=2.5cm,top=2cm,bottom=3cm]{geometry}
\usepackage[font={small,it}]{caption}	
\title{\textsc{What is to Be Gained by Ensemble Models in Analysis of Spectroscopic Data?}}

\author{
    Katarina Domijan
   \\
    Department of Mathematics and Statistics \\
    Maynooth University \\
  Ireland \\
  \texttt{\href{mailto:katarina.domijan@mu.ie}{\nolinkurl{katarina.domijan@mu.ie}}} \\
  }
\date{}                     
\begin{document}

\maketitle

\begin{abstract}
An empirical study was carried out to compare different implementations
of ensemble models aimed at improving prediction in spectroscopic data.
A wide range of candidate models were fitted to benchmark datasets from
regression and classification settings. A statistical analysis using
linear mixed model was carried out on prediction performance criteria
resulting from model fits over random splits of the data. The results
showed that the ensemble classifiers were able to consistently
outperform candidate models in our application.
\end{abstract}


\section{Introduction}\label{sec:intro}

Vibrational spectroscopic techniques, including near-infrared (NIR),
mid-infrared (MIR), and Raman, use the effect of light to provide
information about the constituents of a sample. These low cost, rapid
and noninvasive techniques are widely and routinely used in many
application domains. Prediction in spectroscopic data is a topic of
major interest in chemometric literature, see for example
\cite{Frizzarin2021, Frizzarin2021c, Singh2019}. Numerous advances in
statistical machine learning model methodology in the past few decades
offer the potential to improve prediction performance over the
well-established partial least squares (PLS) approach. Comparative
analyses of algorithm prediction ability for spectroscopic data have
shown that PLS variants perform strongly
\cite{Frizzarin2021c,Singh2019}, but that there isn't a single model
that will outperform others in all settings. Results are context
dependent and specifying only a single model is not the optimal strategy to generate predictions.

Ensemble methods is an umbrella term for model combination techniques to improve prediction accuracy. Ensembles fit multiple \textit{candidate} machine learning models and combine their predictions in a \textit{meta-learner}, a second layer model. The idea is that the new ensemble model is more robust as the individual weaknesses and biases of candidate models are offset by the strengths of other candidates.

In this article, an empirical study was carried out with the aim of
evaluating to what extent ensemble models can improve prediction from
individual candidate models for spectroscopic data. For this purpose,
datasets from two chemometric data analysis competitions
\cite{Frizzarin2021b, frizzarin2023} were used. The two challenges were
to build better calibration models for prediction of milk quality traits
and animal diet from mid-infrared (MIR) spectra. The prediction
challenges represent statistical learning tasks in both regression and
classification settings.

Many different statistical modelling approaches have been employed for
challenges in \cite{Frizzarin2021b} and \cite{frizzarin2023}. The papers
outline different pre-processing steps taken by the participants
including outlier removal, variable selection and variable
transformation. Participants also employed different statistical
learning algorithms for prediction. The papers studied the outcomes in
order to investigate robustness of the findings and discussed various
insights from the analyses. Therefore, the datasets are good candidates
for benchmarks of further studies of comparative performance of
algorithms such as the one presented in this paper, with the caveat that the performance results cannot be directly compared due to well known issues of cross-validation bias arising from unsupervised preprocessing \citep{moscovich22}.

Main approaches for combining models into ensembles fall under the
following categories: model averaging and majority voting, ensemble
stacking, bagging and boosting. The latter two, involve fitting multiple
models using the same algorithm (weak-learner) either to random
sub-samples of the dataset or sequentially. In the present study, Random
Forests (RF) \cite{breiman01} and Generalized Boosted Regression Model
(GBM)\cite{gbmpaper} with decision trees \cite{breiman1984} as
weak-learners were considered. In contrast, model averaging and majority
voting can create an ensemble from a collection of diverse models by
simply taking the average or the majority vote of the test set
predictions of candidate models. Stacking ensembles or super-learners
\cite{Stone74,Wolpert1992,Breiman1996,VanDerLaan07} are a more
sophisticated framework where higher weights can be given to better
performing algorithms in the ensemble. As a result, the final prediction
is less sensitive to the composition of the candidate model library.
Various models can be used to blend the predictions in the meta-learner.
Regularised regression models are commonly employed as candidate model
predictions tend to be correlated. Over-fitting is avoided by
cross-validation.
\cite{Breiman1996, Leblanc1996, VanDerLaan07, Clarke03} studied some
theoretical properties of stacking.

\cite{Breiman1996} and \cite{Clarke03} point out that model averaging is
likely to be more useful when candidate models are more dissimilar,
however, this heuristic is not supported by rigorous theory. In this
study, a wide range of diverse candidate models were fitted and
different implementations of ensemble models to combine their
predictions were compared. The algorithms were trained over random
splits of the data into training and testing sets and a linear mixed
model was fitted to the resulting datasets of prediction performance
measures. The mixed models accounted for within-random split nature of
the experimental setting by including random effects for random split
id. The statistical modelling of algorithm performance metrics is a
secondary novel aspect of this study. The statistical analysis of the
performance metrics showed that the ensemble classifiers were able to
consistently outperform candidate models over different datasets and
random splits.

This paper is organized as follows. The datasets and the methodology are
described in Section \ref{sec:methods}. This includes the candidate
models, cross-validation set up, ensemble approaches for combining the
candidate predictions, performance evaluation metrics and the
statistical analysis of prediction results using a linear mixed model.
In Section \ref{sec:results} the prediction results of various models
are presented. The findings are discussed and concluding statements are
made in Section \ref{sec:discussion}.

\section{Methods}\label{sec:methods}

\subsection{Data Description}

Datasets used in this study originated from two data analysis challenges
organised by the Vistamilk SFI Research Centre as part of International
Workshop on Spectroscopy and Chemometrics 2021 \cite{Frizzarin2021b} and
2022 \cite{frizzarin2023}. Both competitions involved a supervised
learning task from mid-infrared (MIR) spectra of cow milk samples. The
first competition (2021) involved prediction of continuous traits
(regression) and the second (2022) prediction of categorical traits
(classification). From now on in this article we shall refer to the two
competitions as \emph{regression challenge} and \emph{classification
challenge}.

\subsubsection{Regression Challenge Dataset.}

\begin{figure}[h!]
\centering
\includegraphics[width=80mm,scale=0.8]{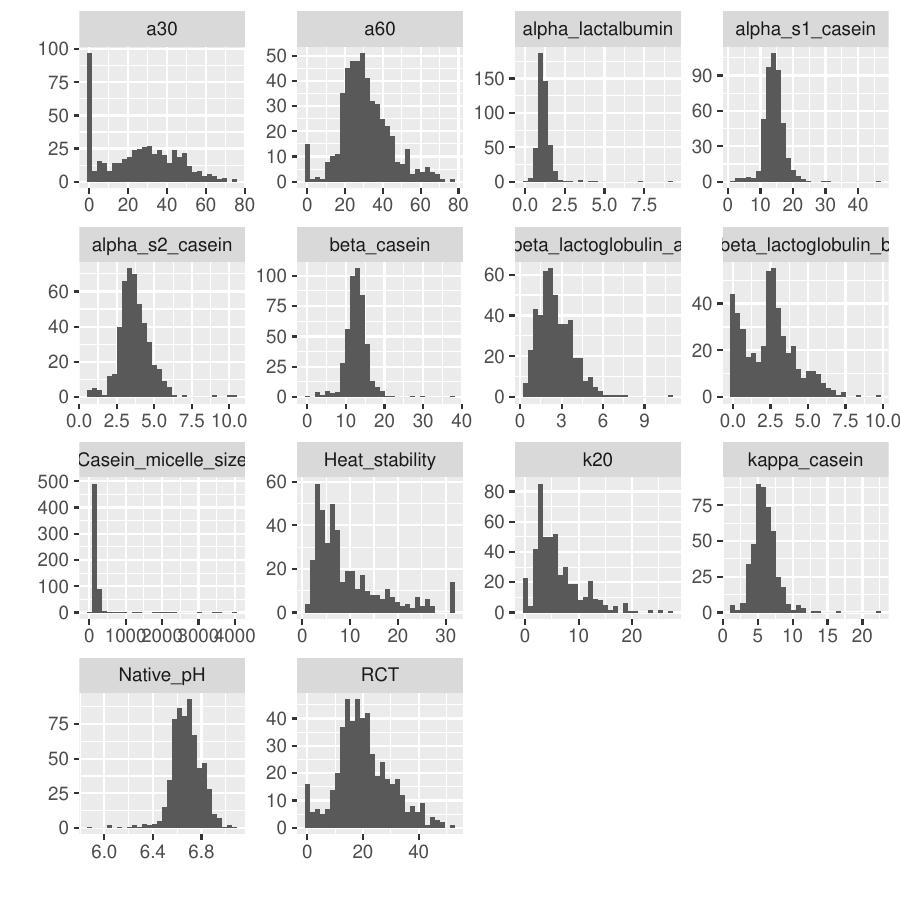}
\caption{Distributions of the fourteen traits in the regression challenge dataset.}\label{fig:traits}
\end{figure}

The dataset comprised mid-infrared (MIR) spectra from milk samples of
622 individual cows. The challenge was to predict fourteen known
technological and milk protein composition traits from spectral
measurements. The traits included rennet coagulation time (RCT),
curd-firming time (k20), curd firmness at 30 and 60 min (a30, a60),
casein micelle size (CMS), pH, heat stability, \(\alpha_{S1}\)-CN,
\(\alpha_{S2}\)-CN, \(\beta\)-CN,\(\kappa\)-CN,
\(\alpha\)-LA,\(\beta\)-LG A, and \(\beta\)-LG B. Figure
\ref{fig:traits} presents histograms of the 14 traits, showing their
distributions. Traits with right-skewed distributions were
log-transformed and all traits were centered and scaled. No outliers
were removed. The dataset is described in detail in \cite{visentin2016},
\cite{mcdermott2016} and \cite{Frizzarin2021c}.

\subsubsection{Classification Challenge Dataset.}

\begin{figure}
\centering
\includegraphics[width=80mm,scale=0.8]{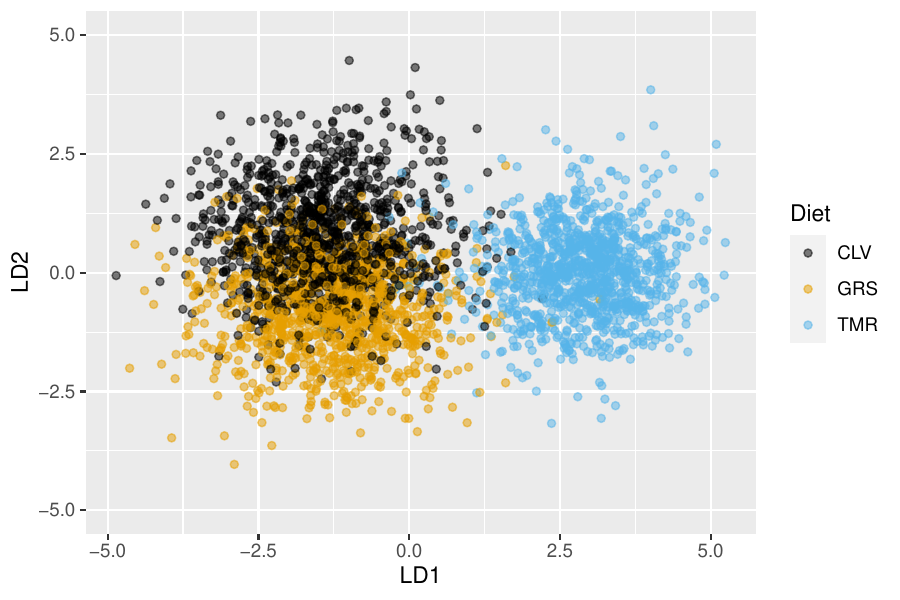}
\caption{Scatter plot of the MIR spectra from the classification challenge dataset projected on the space spanned by the two latent variables associated with the discriminant functions for milk from cows fed grass (GRS), clover (CLV) and nutional mix (TMR).}\label{fig:diets}
\end{figure}

The dataset comprised 3,275 milk MIR spectra of animals fed one of diet
regiments: grass-only (GRS), grass-white clover (CLV) and nutional mix
containing grass silage, maize silage, and concentrates (TMR)
\cite{ocallaghan2016,Frizzarin2021,frizzarin2023}. The task was to
predict animal diet type from MIR spectral measurements of the milk
samples. The dataset was class balanced with 1,094 spectra for GRS,
1,120 spectra for CLV and 1,061 spectra for TMR. Figure \ref{fig:diets}
shows the MIR spectra projected to the latent space spanned by the 2
discriminant functions obtained by liner discriminant analysis (LDA).
The colouring of the points displays the diet regimen of the animals
that the sample had been collected from. The plot illustrates that the
spectra from samples from the two pasture classes (GRS and CLV) tend to
be similar and are harder to discriminate between, compared to the TMR
diet which leads to different milk composition.

\subsubsection{MIR Spectra.}

\begin{figure*}
\centering
\includegraphics[width=75mm,height=48mm]{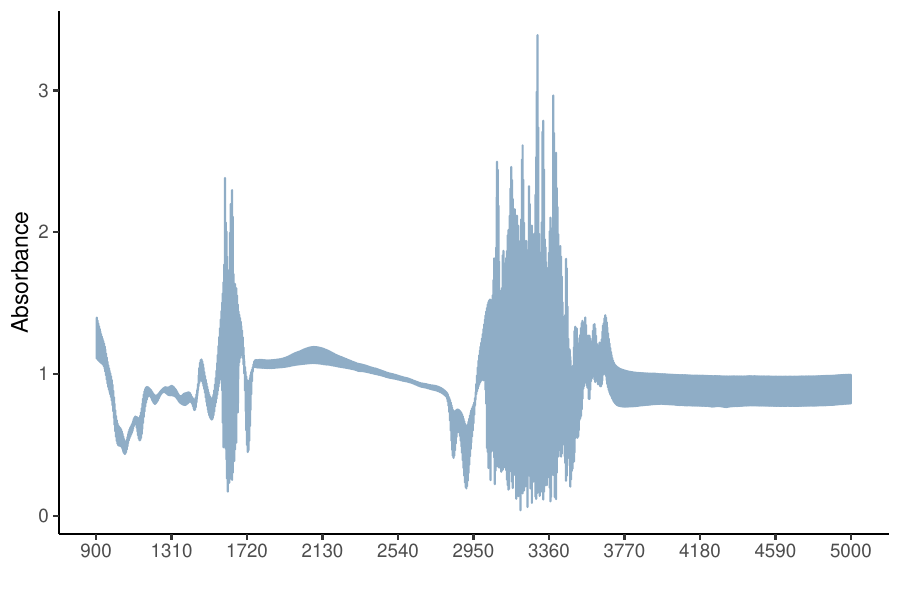}
\includegraphics[width=75mm,height=48mm]{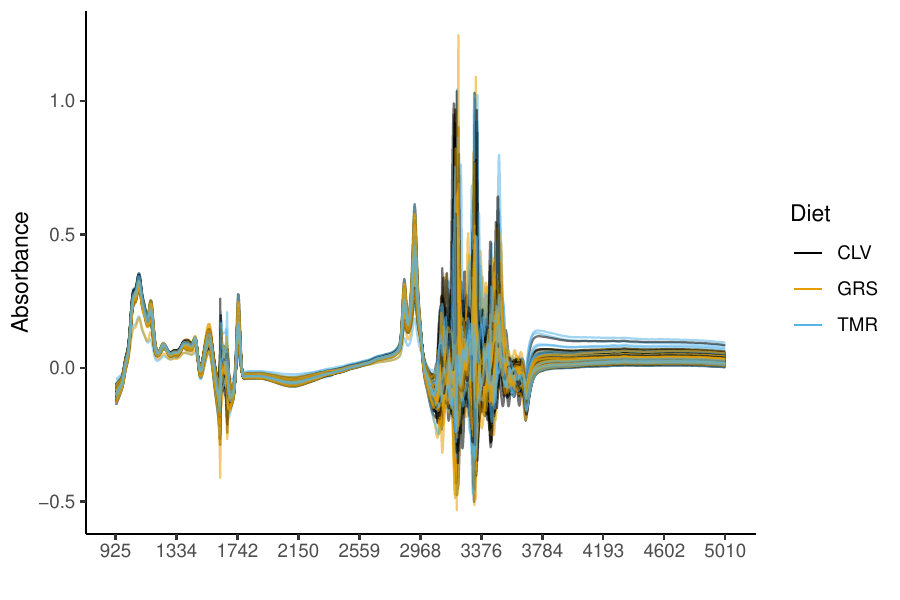}
\caption{Subsets of the of MIR spectra from the regression challenge dataset (left) and classification challenge dataset (right). In the supervised classification task (right), the colours correspond to the diet regimen of the animal whose sample is plotted.}\label{fig:class}
\end{figure*}

The MIR spectra in both datasets were recorded as transmittance values
over a sequence of 1,060 wavelengths in the mid-infrared region (regions
from 900 cm\(^{-1}\)- 5,000 cm\(^{-1}\) and 925 cm\(^{-1}\) - 5,010
cm\(^{-1}\) for regression and classification respectively). In both
datasets, the wavelength transmittance values were transformed to
absorbance by taking log\(_{10}\) of the reciprocal of the transmittance
value. High-noise-level regions: 1,710 - 1,600 cm\(^{-1}\), 3,690 -2,990
cm\(^{-1}\), \textgreater{} 3,822 cm (regression) and 1720 -
1592cm\(^{-1}\), 3698 - 2996 cm\(^{-1}\), \(>\) 3,818 cm\(^{-1}\)
(classification) were removed from each spectrum. This left 531 and 533
wavelengths respectively for the analyses. Figure \ref{fig:class} shows
subsets of the observed mid-infrared milk spectra from the two datasets.
The classification challenge spectra are coloured by the diet regimen of
the animal from whom the sample was taken.

For the classification task, Fisher score (ratio of between to within
diet group variance) and genetic algorithm (GA) \cite{Holland2019} were
used to further reduce the number of input wavelengths. These methods
are implemented in R libraries \cite{bkpc} and \cite{genalg}
respectively and the workflow is described in \cite{frizzarin2023}.

The datasets from both challenges involved additional known structure
due to sample collection methodology: regression challenge dataset
contained samples from different research herds that were collected at
evening or morning milking and the classification cohallenge dataset
contained samples of repeated measures on same animals collected at
different years. As this information was not made available during the
competitions, it was not used in the current study to improve prediction
models.

\subsection{Statistical Analyses}

All the analyses were conducted using the R statistical software
\cite{R}. The code is available at
\url{https://github.com/domijan/ensemblepapercode}.

\subsubsection{Cross-validation for Regression
Challenge.}

To compare the performance of different statistical machine learning
models, the data were randomly split into 50 training and testing sets.
At each random split, candidate models were fit to the training sets.
Further ten-fold cross-validation of the training sets was used for
algorithm tuning. Test set predictions in each random split were
evaluated using root mean squared error (RMSE). The data splitting and
model training and evaluating were performed separately for each of the
14 traits.

\subsubsection{Cross-validation for Classification Challenge.}

Data were randomly split into 10 training and testing sets stratified by
diet. The number or random splits was reduced in comparison to the
regression challenge to reflect the larger sample size. Candidate models
were fit to the training sets with further ten-fold cross-validation for
algorithm tuning. As the dataset is class balanced, the predictions for
the test sets in each random split were evaluated using accuracy (ACC)
which was defined as the proportion of correctly classified samples
divided by the total number of samples.

\subsubsection{Candidate Models.}

The list of candidate models used in the stacking ensembles for the
regression and classification challenge is given in Table
\ref{tab:table}.

\begin{table}
\tiny
 \caption{Candidate Algorithms}
   \centering
  \begin{tabular}{p{75mm}|l|l}
    \toprule
    Algorithm    & Implementation  & Challenge  \\
\hline
Partial Least Squares (PLS) \cite{garthwaite1994} &  \cite{pls}  & regression, classification \\
Linear Regression with PCA (LM + PCA)    & base R &regression \\
Linear Discriminant Analysis (LDA) &  \cite{MASS}& classification\\
Least Absolute Shrinkage and Selection Operator (LASSO)
  \cite{Tibshirani1996}    &  \cite{glmnet}  & regression, classification   \\
Elastic Net (EN) \cite{Zou2005}   &  \cite{glmnet}  & regression, classification   \\
Random Forest (RF) \cite{breiman01}    &  \cite{ranger}  & regression, classification   \\
Random Forest with Variable Selection (RF+VS) \cite{Wundervald2020GeneralizingGP}  & \cite{ranger}   & regression \\
Linear Regression with kPCA \cite{kpca} (LM +  kPCA)    &   \cite{bkpc}   & regression  \\
Random Forest with kPCA (RF + kPCA)   &   \cite{ranger} ,   \cite{bkpc}   & regression \\
Random Forest with Gaussian kernel (RF + kernel)   &   \cite{ranger} ,   \cite{bkpc}   & regression \\
Support Vector Machine (SVM) \cite{vapnik98}  &   \cite{kernlab}  & regression, classification  \\
Bayesian Additive Regression Trees (BART) \cite{bart}   &   \cite{bartMachine}   & regression  \\
Generalized Boosted Regression Model (GBM)\cite{gbmpaper}    & \cite{gbm}  & regression   \\
Bayesian Neural Net (BNN)   & \cite{brnn}  & regression   \\
Neural Net (NNET) & \cite{MASS} & classification\\
Projection Pursuit Regression (PPR) \cite{ppr}   & base R & regression  \\
\bottomrule
\end{tabular}
\label{tab:table}
\end{table}

Based on their characteristics, the selected candidate models can be
grouped into the following classes: dimension reduction methods,
regularised regression methods, kernel methods, neural networks and tree
based ensemble methods. The classes of models were selected with
consideration of their applicability to spectroscopic data and with the
aim of increasing the diversity of the models in the library. A very
brief outline of each class of methods is given below, with some
justification of their choice. A detailed review of these techniques is
given in \cite{hastie01}.

\paragraph{Dimension Reduction Methods.}

Spectroscopic data typically comprise a large number of highly serially
correlated features. Linear models that involve dimension reduction or
variable selection tend to perform well for datasets of this structure.
Dimension reduction methods produce a lower dimensional representation
of the data using new variables that are derived from linear
combinations of the original wavelengths. Partial least squares (PLS)
\cite{garthwaite1994} for regression and classification is the most
well-known and established approach of this type for spectroscopic data.
Other approaches tend to combine principal component analysis (PCA)
\cite{pca} with a linear model such as linear regression in the
regression tasks and logistic regression or linear discriminant analysis
(LDA) in classification tasks, see for example \cite{odwyer2021}. In
addition to being a popular algorithm for linear discrimination, LDA
also provides a lower dimensional representation of the spectra.
Projection pursuit regression (PPR) \cite{ppr} is another model that
combines dimension reduction with a linear model for prediction.

\paragraph{Regularized Regression Methods.}

Regularized regression methods such as Least Absolute Shrinkage and
Selection Operator (LASSO)\cite{Tibshirani1996} and Elastic Net (EN)
\cite{Zou2005} include a penalty term in the loss function that has the
effect of shrinking some regression coefficients to zero. This leads to
variable selection as corresponding wavelengths are removed from the
model. The resulting models are more parsimonious and interpretable in
contrast with the dimension reduction methods where all the wavelengths
are utilised in obtaining the lower dimensional representations of the
data. Regularized regression methods can be formulated for the
classification tasks using logistic regression. They perform well in
highly dimensional datasets with many correlated variables, such as the
application in this study, however, the strength of the penalisation is
very sensitive to tuning parameters and can strongly affect the fit.

\paragraph{Kernel Methods.}

Kernel methods such as Support Vector Machine (SVM) \cite{vapnik98} work
well with spectroscopic data as they function through a matrix of
pairwise distances of observations across the feature space, rather than
the data coordinates. Thus the high dimensionality of the dataset does
not increase the complexity of the algorithms and the high correlation
does not pose estimation problems. However, if only a small number of
wavelengths are useful for the statistical learning task, the additional
spurious wavelengths would have the effect of masking the signal in the
distance (kernel) matrix, so it can be useful to combine kernel models
with a wavelength selection pre-processing step. Different choices of
the kernel matrix implicitly map the data to a higher dimensional space
where the linear algorithm is applied, therefore elegantly inducing
nonlinearity in the models. Model performance can be highly sensitive to the kernel matrix hyper-parameters which require careful tuning.

\paragraph{Neural Networks.}

Neural network (NN) models involve constructing new features from linear
combinations of the original wavelengths and combining the derived
features though a second layer linear regression algorithm. Multiple
intermediary layers are constructed, thus producing very flexible,
nonlinear models. NNs are particularly successful in domains where the
features of the datasets have complicated hierarchical structures such
as image or text data. The dataset sizes in this study are not ideal for
training deep networks so the architectures were limited to 2-layer
networks. Furthermore, deep architectures are complex to tune and
computationally burdensome, so are not easily included as candidate
models for ensembles.

\paragraph{Tree-based Ensembles.}

Tree-based methods fit nonlinear models to the data, but using a
different approach from neural networks and kernel based models. A
decision tree is a simple model that recursively segments the prediction
space into non-overlapping regions. The prediction at each sub-region is
the average of the trait measurements (regression) or the most commonly
occurring class (classification). Random Forests (RFs)\cite{breiman01}
fit multiple trees on bootstrapped samples of the dataset and combine
their predictions into a consesus prediction using majority vote
(classification) or averaging (regression). Only a subset of randomly
selected wavelengths is used for constructing each tree in order to
diversify them. Boosting approaches such as Generalized Boosted
Regression Model (GBM)\cite{gbmpaper} fit trees sequentially to the
residuals obtained from previously fitted tree. Both are ensemble
methods as they borrow strength from a collection of weak-learner
algorithms (trees).

\subsubsection{Meta-Learners.}

To train stacking algorithms, a second layer model (meta-learner) was used to combine predictions from candidate models at each random split. To avoid over-fitting, further ten-fold cross-validation on the training sets within each random split was needed. Meta-learners were trained on the out-of-fold predictions of the candidate models and the resulting coefficients were used to combine predictions in the corresponding test sets from the same random split. Pseudo-code for stacking algorithm implemented for regression challenge is given in Algorithm \ref{alg1}. For the classification challenge, the R based implementation of stacking in package \texttt{stacks} \cite{stacks} was used. The default \texttt{stacks} implementation fits a selection of candidate models over pre-determined grids of tuning parameters and blends the predictions of the candidate models from all the tuning parameter settings in the grids. In contrast, the study presented in this paper used the implementation that combined predictions of tuned models. Note that many variants and implementations of stacking exist. For accessible primers, see, for example, \cite{Wolfinger2017SAS2017SE} and \cite{naimi2018}.

Standard meta-learner choices are linear regression with a constraint for non-negative coefficients (ens\_nonneg) or with shrinkage penalties such as LASSO (ens\_LASSO). For the regression challenge we also considered unconstrained linear regression (ens\_LM) and a random forest (ens\_RF). For the classification challenge we used logistic regression with non-negative restriction for the coefficients (ens\_nonneg) which is the default meta-learner in \texttt{stacks}.

For comparison with stacking ensembles and tree based ensemble models, model averaging (regression challenge) and majority vote (classification challenge) were also considered. These ensembles simply combined the test set predictions from the candidate models in each random split. The additional cross-validation layer used for stacking was not needed.

\begin{algorithm}
\caption{Stacking ensemble algorithm pseudo-code}\label{alg1}
\begin{algorithmic}[1]
\State Randomly split the dataset into $N$ training and testing sets.
\For{$i \leftarrow 1$ to $N$}
    \State Use 10-fold cross-validation of training set $i$ to train all candidate models and save their out-of-fold predictions. A further cross-validatory layer may be required for tuning parameters of candidate models.
    \State Combine out-of-fold predictions of candidate models into a dataset and train a meta-learner model.
    \State Train all candidate models on the full training set $i$ and obtain predictions for the corresponding test set $i$.
    \State Use meta-learner coefficients estimated from 4 to combine test set $i$ predictions obtained in 5.
\EndFor
\end{algorithmic}
\end{algorithm}

\subsubsection{Performance Evaluation with the Linear Mixed Model
(LME).}

Statistical analysis of the data of the prediction performance metrics
arising from the study was carried out to compare the predictive ability
of different algorithms and the ensemble models. In order to correctly
account for the experimental design of our study (algorithms were
trained within random splits) when comparing average performance
criteria (RMSE, ACC), a linear mixed effects (LME) model \cite{lme4} was
fit.

\begin{figure}
\centering
\includegraphics[width=80mm,scale=0.8]{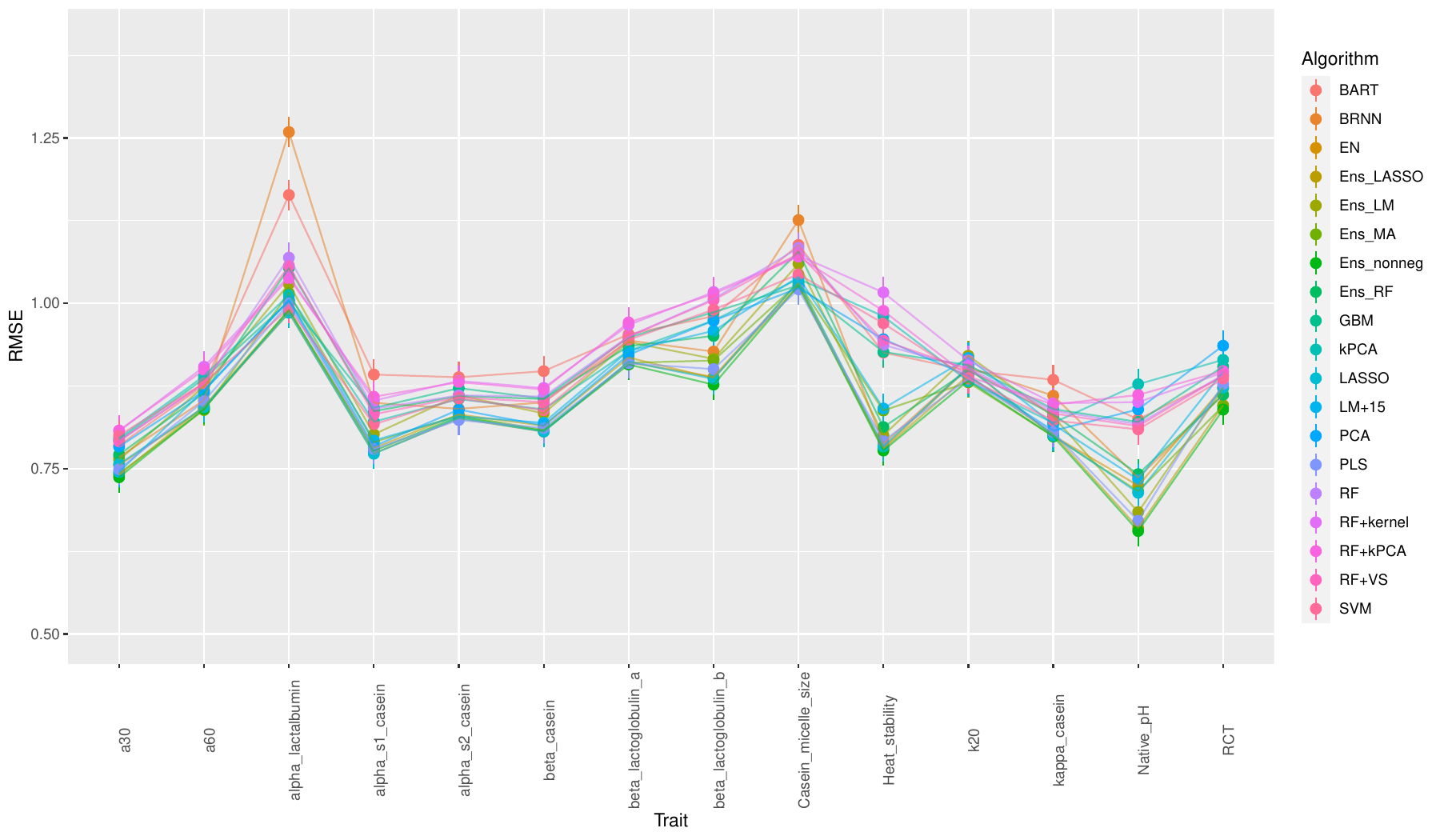}
\caption{Regression dataset: LME estimates of mean RMSE with 95\% CI. Overlapping intervals indicate where the difference is not statistically significant. The interaction plot shows that for most traits (x-axis), the stacking ensemble with non-negative constraint on the
coefficients (Ens\_nonneg, solid green line), gives lowest mean RMSE.}\label{fig:fig5a}
\end{figure}

\begin{figure}
\centering
\includegraphics[width=80mm,scale=0.8]{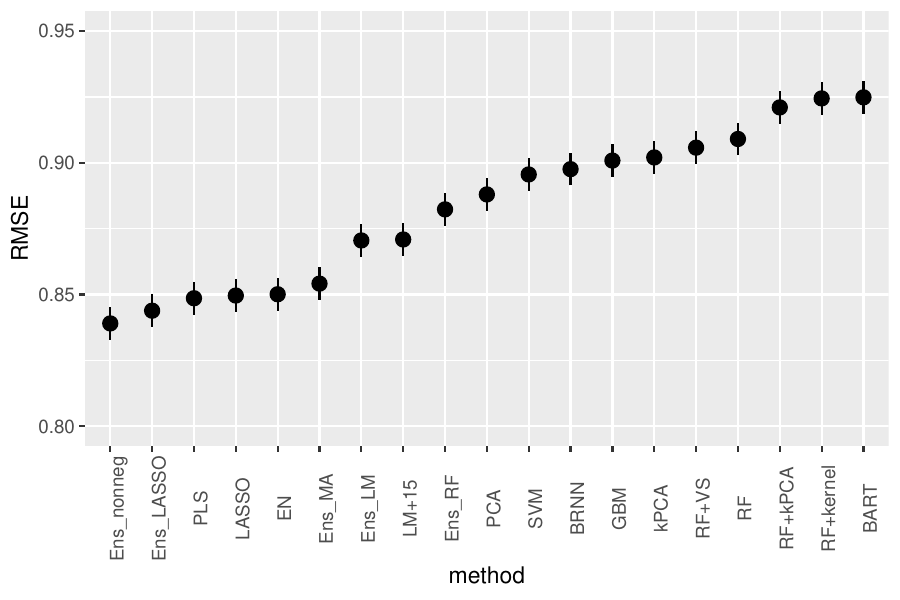}
\caption{Regression dataset: LME model estimates of mean RMSE over all random splits and traits with 95\% CIs. Overlapping intervals indicate where the difference is not
statistically significant. The plot shows that the stacking ensemble with the non-negative coefficients gets the lowest average RMSE over all random splits and traits. Second best is stacking ensemble with lasso (Ens\_LASSO) and the top candidate algorithms are PLS, LASSO and EN. These candidate models outperform the model averaging ensemble (Ens\_MA).}\label{fig:fig5b}
\end{figure}

Evaluation of the the test set predictions from random splits of the
data yielded two new datasets of RMSE or ACC for each of the candidate
and ensemble models at each random split. In the regression dataset,
RMSE was obtained from modeling the 14 traits which yielded a dataset of
13,300 data points (14 traits \(\times\) 50 random splits \(\times\) 19
models). In the classification setting, the resulting dataset comprised
ACC for 10 random splits \(\times\) 9 models, giving 90 data points. To
compare the effect of candidate algorithm choice on RMSE or ACC, an LME
was fit to the two datasets. In the classification challenge dataset,
algorithm type was treated as a fixed effect and in the regression
challenge dataset, trait and trait-algorithm interaction were
additionally included as fixed effects. The LMEs accounted for
within-random split nature of the experimental setting by including
random effects for random split id. Post-hoc tests were used to find
significant differences between mean RMSE/ACC of different models. R
based package \texttt{effects} \cite{effects} was used to obtain 95\%
CIs for the effects in Figures \ref{fig:fig5a}, \ref{fig:fig5b} and
\ref{fig:fig6b}.

\section{Results}\label{sec:results}

\begin{figure}
\centering
\includegraphics[width=90mm,scale=0.8]{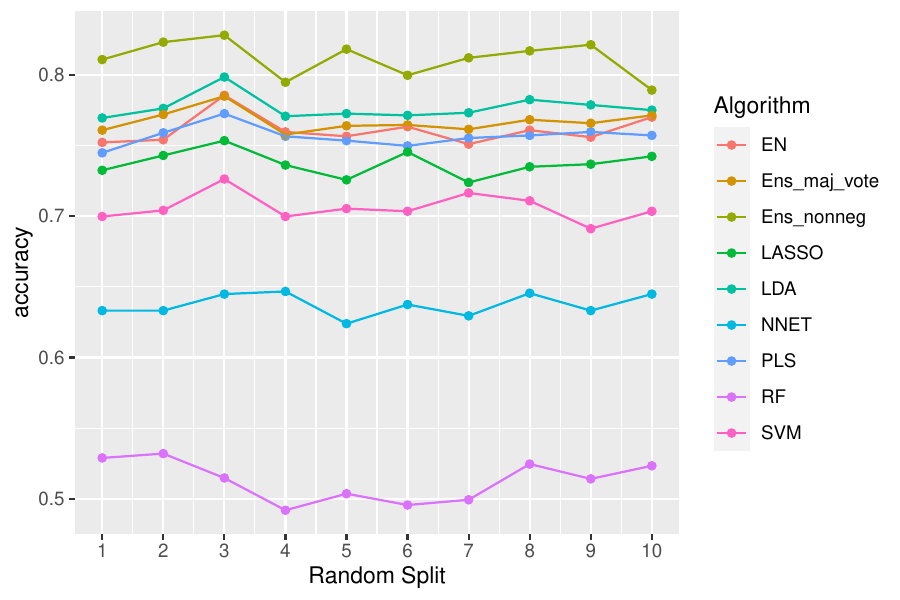}
\caption{Classification dataset: ACC of algorithms over 10 random
splits. Stacking ensemble with non-negative constraint on the
coefficients (Ens\_nonneg) achieves accuracy of \textasciitilde81\%,
which is consistently higher than the best performing candidate model (LDA) with ACC \textasciitilde78\%. Majority vote ensemble gets lower ACC than LDA at all random splits.}\label{fig:fig6a}
\end{figure}

\begin{figure}
\centering
\includegraphics[width=80mm,scale=0.8]{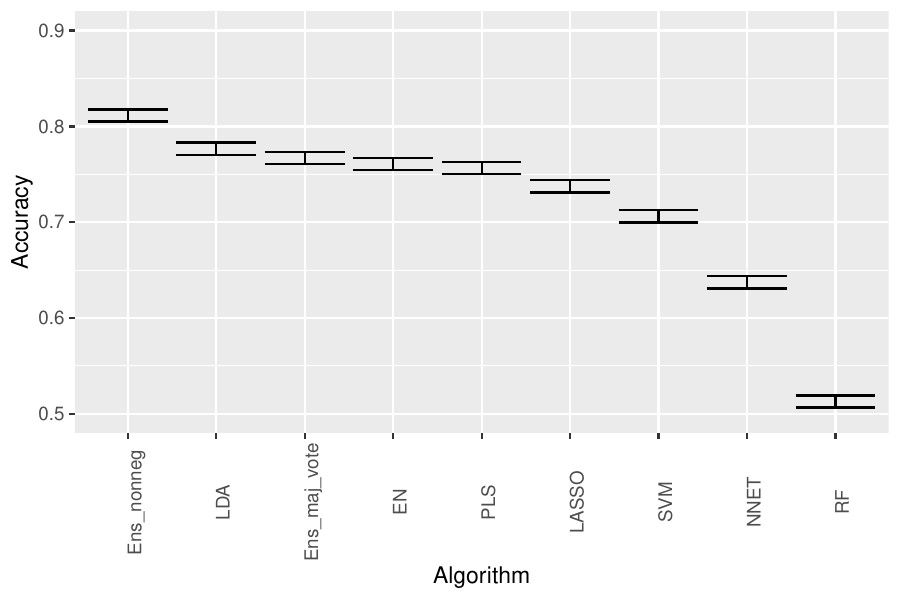}
\caption{Classification dataset: LME model estimates of mean ACC with 95\% CIs by algorithm. The plot shows that the stacking ensemble with the non-negative coefficients gets the highest ACC on average and this is a statistically significant (non-overlapping CIs) improvement on the performance of the best candidate model (LDA). LDA  outperforms majority vote ensemble.}\label{fig:fig6b}
\end{figure}

\begin{figure}
\centering
\includegraphics[width=80mm,scale=0.8]{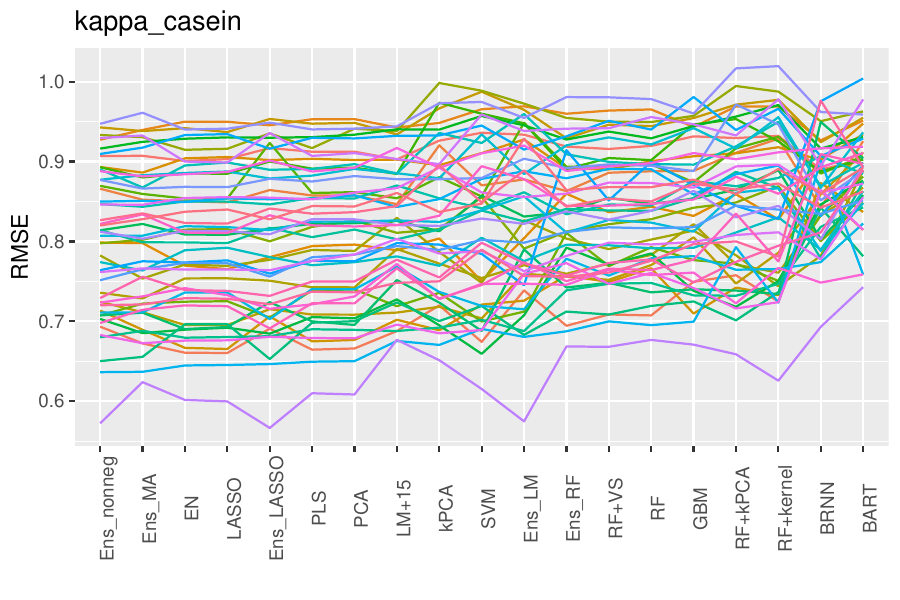}
\caption{Regression dataset: RMSE of algorithms over 50 random
splits for one of the traits $\kappa$-casein.}\label{fig:fig7a}
\end{figure}

\subsection{Regression Challenge}

There is a significant difference in the average RMSE for different
algorithms and the significant algorithm-trait interaction (p-value
\(<\) 0.05) in the regression challenge dataset, indicating some
algorithms performed significantly better than others at predicting
different traits. This can be seen by non-parallel and intersecting
lines in the interaction plot in Figure \ref{fig:fig5a}. Non-overlapping
95\% CIs for the mean RMSE indicate a significant difference in the
prediction performance.

Stacking ensemble with non-negative constraint on the coefficients
(Ens\_nonneg) consistently performed at least as well as the best
candidate model in all traits, which can be seen in Figure
\ref{fig:fig5a} where Ens\_nonneg achieves lowest average RMSE at all
traits.

Figure \ref{fig:fig5b} shows the estimated average RMSE from the LME
model. The stacking ensemble with non-negative coefficients gets the
lowest average RMSE over all random splits and traits. Second best is
stacking ensemble with lasso (Ens\_LASSO) and the difference in average
RMSE between the two is not statistically significant. These models
performed significantly better than other ensemble approaches for
blending predictions including linear regression without constraints
(Ens\_LM) and random forest (Ens\_RF). The top candidate algorithms were
PLSR, LASSO and EN. These candidate models get lower average RMSE than
the model averaging ensemble (Ens\_MA).

\subsection{Classification Challenge}

Figure \ref{fig:fig6a} shows the ACC of all algorithms over 10 random
splits in the classification dataset. Stacking ensemble with
non-negative constraint on the coefficients (Ens\_nonneg) achieves
accuracy of \textasciitilde81\%, which is consistently higher than the
best performing candidate model (LDA) with ACC \textasciitilde78\%.
Majority vote ensemble gets lower ACC than LDA at all random splits.

Figure \ref{fig:fig6b} shows the LME model estimates of mean ACC with
95\% CI by algorithm. Overlapping CIs indicate where the difference is
not statistically significant. The stacking ensemble with non-negative
coefficients (Ens\_nonneg) gets the highest ACC on average and this is a
statistically significant improvement on the performance of the best
candidate model (LDA). LDA outperforms majority vote
ensemble model. As in the regression setting, LASSO, EN and PLS
outperformed nonlinear models (SVM, RF, NNET).

The regression challenge dataset had a relatively small sample size and
in this setting the variability in prediction accuracy due to
cross-validatory splits of the data was much larger rather than
variability in accuracy of candidate models in all fourteen traits. This
is illustrated in Figure \ref{fig:fig7a} which plots RMSE from different
algorithms in the test sets from random splits for one of the traits
\(\kappa\)-casein. Plots for other traits (not shown) follow similar
pattern. Furthermore, the estimated standard deviation between the
random splits in the LME model was twice the estimated residual standard
deviation. In the classification challenge dataset which comprised much
larger sample size, the LME model estimated the standard deviation
between the random splits as half of the estimated residual standard
deviation. In this example, the variability in accuracy of different
algorithms was larger than the variability in ACC due to random splits,
as can be seen in Figure \ref{fig:fig6a}.

\section{Conclusions}\label{sec:discussion}

Stacking ensembles with a rich set of candidate models are a versatile
tool that researchers can use to improve prediction. The statistical
analysis showed that they can consistently improve on the performance of the best candidate models over different datasets and random splits. In the regression dataset, mean RMSE decreased from 0.85 to 0.84, and in the classification dataset mean ACC increased from 0.78 to 0.81. The
improvement in performance is modest, however these models have another
advantage. Comparative analyses of algorithms in terms of their
prediction ability for spectroscopic data have shown that while well
established methods like PLS perform strongly, there isn't a single
model that will outperform others in all settings. Results are context
specific and can depend on the structure of the dataset at hand \cite{Frizzarin2021c}, \cite{Singh2019}. Therefore, choosing a single model to
generate predictions is unlikely to be the optimal approach, and many
potential models are available. Stacking ensembles offer an elegant way
of combining predictions of different candidate models.

Since stacking ensemble algorithms use weighted combinations of
candidate models to improve performance, the intuition is that
increasing the diversity of candidate predictions is likely to increase
the chance that ensemble stacking can improve on the predictions of a
single strong candidate model. It is worth noting that if the predictions of the candidate models do not show much variation, the aggregation using stacking will add complexity without benefit.

In our stacking ensemble model implementations, candidate models of
differing characteristics were selected. LASSO and EN use constrained
optimisation to reduce the number of wavelengths that is input into
linear models, dimension reduction methods project data to lower
dimensional spaces (PLS, LDA, PCA + LM) but retain all wavelengths for
computation, regression and classification algorithms that employ
different ways of inducing nonlinearity in the model e.g.~tree based
methods (RF, BART, GBM), kernel based methods (SVM) and neural networks.
The tree-based ensemble methods (RF, BART) were the weakest candidate
models in terms of their predictive ability for these datasets, whereas
linear models like PLS and LDA and regularized regression methods
(LASSO, EN) were the strongest candidate models. Spectroscopic data
present a specific structure of very high dimension and strong serial
correlation between wavelengths, so these classes of models are likely
to show a different pattern of performance in different applications.

While there was some variability in algorithm performance for different
traits, the LME model showed the stacking ensembles significantly outperformed model averaging (Ens\_MA)
and majority voting (Ens\_maj\_vote) ensembles in our application. These are poor approaches to combining model predictions and, on average, performed worse than some candidate models.

Stacking ensemble model implementations can increase diversity of
predictions by using different hyper-parameter settings, however we
chose to blend the predictions of tuned models. Other ways of enhancing
diversity of predictions in the literature involve training the models
over different randomly selected subsets of the wavelengths and
generating various versions of the training data such as by bagging.
This was beyond the scope of the current empirical study. Instead,
different choices of meta-learners for blending the predictions were
investigated and found that using a non-negative constraint on the
coefficients of the linear model gave the best results, which is in
agreement with the results of earlier empirical studies
\cite{Breiman1996,Leblanc1996}.

\bibliographystyle{apalike}

\bibliography{rscens}

\begin{thebibliography}{}

\bibitem[Bates et~al., 2015]{lme4}
Bates, D., M{\"a}chler, M., Bolker, B., and Walker, S. (2015).
\newblock Fitting linear mixed-effects models using {lme4}.
\newblock {\em Journal of Statistical Software}, 67(1):1--48.

\bibitem[Breiman, 1996]{Breiman1996}
Breiman, L. (1996).
\newblock Stacked regressions.
\newblock {\em Machine Learning}, 24:49--64.

\bibitem[Breiman, 2001]{breiman01}
Breiman, L. (2001).
\newblock Random forests.
\newblock {\em Machine Learning}, 45(1):5--32.

\bibitem[Breiman et~al., 1984]{breiman1984}
Breiman, L., Friedman, J., Stone, C., and Olshen, R. (1984).
\newblock {\em Classification and Regression Trees}.
\newblock Taylor \& Francis.

\bibitem[Chipman et~al., 2010]{bart}
Chipman, H.~A., George, E.~I., and McCulloch, R.~E. (2010).
\newblock {BART: Bayesian additive regression trees}.
\newblock {\em The Annals of Applied Statistics}, 4(1):266 -- 298.

\bibitem[Clarke, 2003]{Clarke03}
Clarke, B. (2003).
\newblock Comparing bayes model averaging and stacking when model approximation
  error cannot be ignored.
\newblock {\em J. Mach. Learn. Res.}, 4(null):683–712.

\bibitem[Couch and Kuhn, 2022]{stacks}
Couch, S. and Kuhn, M. (2022).
\newblock {\em stacks: Tidy Model Stacking}.
\newblock R package version 0.2.3.

\bibitem[Domijan, 2018]{bkpc}
Domijan, K. (2018).
\newblock {\em BKPC: Bayesian Kernel Projection Classifier}.
\newblock R package version 1.0.1.

\bibitem[Fox and Weisberg, 2018]{effects}
Fox, J. and Weisberg, S. (2018).
\newblock Visualizing fit and lack of fit in complex regression models with
  predictor effect plots and partial residuals.
\newblock {\em Journal of Statistical Software}, 87(9):1--27.

\bibitem[Friedman et~al., 2010]{glmnet}
Friedman, J., Hastie, T., and Tibshirani, R. (2010).
\newblock Regularization paths for generalized linear models via coordinate
  descent.
\newblock {\em Journal of Statistical Software}, 33(1):1--22.

\bibitem[Friedman, 2001]{gbmpaper}
Friedman, J.~H. (2001).
\newblock Greedy function approximation: A gradient boosting machine.
\newblock {\em The Annals of Statistics}, 29(5):1189 -- 1232.

\bibitem[Friedman and Stuetzle, 1981]{ppr}
Friedman, J.~H. and Stuetzle, W. (1981).
\newblock Projection pursuit regression.
\newblock {\em Journal of the American Statistical Association},
  76(376):817--823.

\bibitem[Frizzarin et~al., 2021a]{Frizzarin2021b}
Frizzarin, M., Bevilacqua, A., Dhariyal, B., Domijan, K., Ferraccioli, F.,
  Hayes, E., Ifrim, G., Konkolewska, A., Nguyen, T.~L., Mbaka, U., Ranzato, G.,
  Singh, A., Stefanucci, M., and Casa, A. (2021a).
\newblock Mid infrared spectroscopy and milk quality traits: A data analysis
  competition at the international workshop on spectroscopy and chemometrics
  2021.
\newblock {\em Chemometrics and Intelligent Laboratory Systems}, 219.

\bibitem[Frizzarin et~al., 2021b]{Frizzarin2021c}
Frizzarin, M., Gormley, I., Berry, D., Murphy, T., Casa, A., Lynch, A., and
  McParland, S. (2021b).
\newblock Predicting cow milk quality traits from routinely available milk
  spectra using statistical machine learning methods.
\newblock {\em Journal of Dairy Science}, 104(7):7438--7447.

\bibitem[Frizzarin et~al., 2021c]{Frizzarin2021}
Frizzarin, M., O'Callaghan, T.~F., Murphy, T.~B., Hennessy, D., and Casa, A.
  (2021c).
\newblock Application of machine-learning methods to milk mid-infrared spectra
  for discrimination of cow milk from pasture or total mixed ration diets.
\newblock {\em Journal of Dairy Science}, 104(12):12394--12402.

\bibitem[Frizzarin et~al., 2023]{frizzarin2023}
Frizzarin, M., Visentin, G., Ferragina, A., Hayes, E., Bevilacqua, A.,
  Dhariyal, B., Domijan, K., Khan, H., Ifrim, G., Nguyen, T.~L., Meagher, J.,
  Menchetti, L., Singh, A., Whoriskey, S., Williamson, R., Zappaterra, M., and
  Casa, A. (2023).
\newblock Classification of cow diet based on milk mid infrared spectra: A data
  analysis competition at the “international workshop on spectroscopy and
  chemometrics 2022”.
\newblock {\em Chemometrics and Intelligent Laboratory Systems}, 234:104755.

\bibitem[Garthwaite, 1994]{garthwaite1994}
Garthwaite, P.~H. (1994).
\newblock An interpretation of partial least squares.
\newblock {\em Journal of the American Statistical Association},
  89(425):122--127.

\bibitem[Greenwell et~al., 2022]{gbm}
Greenwell, B., Boehmke, B., Cunningham, J., and Developers, G. (2022).
\newblock {\em gbm: Generalized Boosted Regression Models}.
\newblock R package version 2.1.8.1.

\bibitem[Hastie et~al., 2001]{hastie01}
Hastie, T., Tibshirani, R., and Friedman, J. (2001).
\newblock {\em The Elements of Statistical Learning}.
\newblock Springer Series in Statistics. Springer New York Inc., New York, NY,
  USA.

\bibitem[Holland, 2019]{Holland2019}
Holland, J.~H. (2019).
\newblock {\em Adaptation in Natural and Artificial Systems}.

\bibitem[Kapelner and Bleich, 2016]{bartMachine}
Kapelner, A. and Bleich, J. (2016).
\newblock {bartMachine}: Machine learning with {B}ayesian additive regression
  trees.
\newblock {\em Journal of Statistical Software}, 70(4):1--40.

\bibitem[Karatzoglou et~al., 2004]{kernlab}
Karatzoglou, A., Smola, A., Hornik, K., and Zeileis, A. (2004).
\newblock kernlab -- an {S4} package for kernel methods in {R}.
\newblock {\em Journal of Statistical Software}, 11(9):1--20.

\bibitem[Laan et~al., 2007]{VanDerLaan07}
Laan, M. J. V.~D., Polley, E.~C., and Hubbard, A.~E. (2007).
\newblock Super learner.
\newblock {\em Statistical Applications in Genetics and Molecular Biology}, 6.

\bibitem[Leblanc and Tibshirani, 1996]{Leblanc1996}
Leblanc, M. and Tibshirani, R. (1996).
\newblock Combining estimates in regression and classification.
\newblock {\em Journal of the American Statistical Association},
  91(436):1641--1650.

\bibitem[McDermott et~al., 2016]{mcdermott2016}
McDermott, A., Visentin, G., {De Marchi}, M., Berry, D., Fenelon, M.,
  O’Connor, P., Kenny, O., and McParland, S. (2016).
\newblock Prediction of individual milk proteins including free amino acids in
  bovine milk using mid-infrared spectroscopy and their correlations with milk
  processing characteristics.
\newblock {\em Journal of Dairy Science}, 99(4):3171--3182.

\bibitem[Mevik et~al., 2020]{pls}
Mevik, B.-H., Wehrens, R., and Liland, K.~H. (2020).
\newblock {\em pls: Partial Least Squares and Principal Component Regression}.
\newblock R package version 2.7-3.

\bibitem[Moscovich and Rosset, 2022]{moscovich22}
Moscovich, A. and Rosset, S. (2022).
\newblock {On the Cross-Validation Bias due to Unsupervised Preprocessing}.
\newblock {\em Journal of the Royal Statistical Society Series B: Statistical
  Methodology}, 84(4):1474--1502.

\bibitem[Naimi and Balzer, 2018]{naimi2018}
Naimi, A.~I. and Balzer, L.~B. (2018).
\newblock Stacked generalization: an introduction to super learning.
\newblock {\em European journal of epidemiology}, 33(5):459--464.

\bibitem[O’Callaghan et~al., 2016]{ocallaghan2016}
O’Callaghan, T.~F., Hennessy, D., McAuliffe, S., Kilcawley, K.~N.,
  O’Donovan, M., Dillon, P., Ross, R., and Stanton, C. (2016).
\newblock Effect of pasture versus indoor feeding systems on raw milk
  composition and quality over an entire lactation.
\newblock {\em Journal of Dairy Science}, 99(12):9424--9440.

\bibitem[O’Dwyer et~al., 2021]{odwyer2021}
O’Dwyer, K., Domijan, K., Dignam, A., Butler, M., and Hennelly, B.~M. (2021).
\newblock Automated raman micro-spectroscopy of epithelial cell nuclei for
  high-throughput classification.
\newblock {\em Cancers}, 13(19).

\bibitem[Pearson, 1901]{pca}
Pearson, K. (1901).
\newblock Principal components analysis.
\newblock {\em The London, Edinburgh, and Dublin Philosophical Magazine and
  Journal of Science}, 6(2):559.

\bibitem[{R Core Team}, 2020]{R}
{R Core Team} (2020).
\newblock {\em R: A Language and Environment for Statistical Computing}.
\newblock R Foundation for Statistical Computing, Vienna, Austria.

\bibitem[Rodriguez and Gianola, 2022]{brnn}
Rodriguez, P.~P. and Gianola, D. (2022).
\newblock {\em brnn: Bayesian Regularization for Feed-Forward Neural Networks}.
\newblock R package version 0.9.2.

\bibitem[Sch\"olkopf et~al., 1998]{kpca}
Sch\"olkopf, B., Smola, A., and M\"uller, K.-R. (1998).
\newblock Nonlinear component analysis as a kernel eigenvalue problem.
\newblock {\em Neural Computation}, 10(5):1299--1319.

\bibitem[Singh and Domijan, 2019]{Singh2019}
Singh, M. and Domijan, K. (2019).
\newblock Comparison of machine learning models in food authentication studies.
\newblock {\em 30th Irish Signals and Systems Conference, ISSC 2019}.

\bibitem[Stone, 1974]{Stone74}
Stone, M. (1974).
\newblock {C}ross-{V}alidatory {C}hoice and {A}ssessment of {S}tatistical
  {P}redictions.
\newblock {\em Journal of the Royal Statistical Society. Series B
  (Methodological)}, 36(2):111--147.

\bibitem[Tibshirani, 1996]{Tibshirani1996}
Tibshirani, R. (1996).
\newblock Regression shrinkage and selection via the lasso.
\newblock {\em Journal of the Royal Statistical Society: Series B
  (Methodological)}, 58:267--288.

\bibitem[Vapnik, 1998]{vapnik98}
Vapnik, V. (1998).
\newblock {\em Statistical Learning Theory}.
\newblock Wiley-Interscience, New York.

\bibitem[Venables and Ripley, 2002]{MASS}
Venables, W.~N. and Ripley, B.~D. (2002).
\newblock {\em Modern Applied Statistics with S}.
\newblock Springer, New York, fourth edition.
\newblock ISBN 0-387-95457-0.

\bibitem[Visentin et~al., 2016]{visentin2016}
Visentin, G., Penasa, M., Gottardo, P., Cassandro, M., and {De Marchi}, M.
  (2016).
\newblock Predictive ability of mid-infrared spectroscopy for major mineral
  composition and coagulation traits of bovine milk by using the uninformative
  variable selection algorithm.
\newblock {\em Journal of Dairy Science}, 99(10):8137--8145.

\bibitem[Willighagen and Ballings, 2022]{genalg}
Willighagen, E. and Ballings, M. (2022).
\newblock {\em genalg: R Based Genetic Algorithm}.
\newblock R package version 0.2.1.

\bibitem[Wolfinger and Tan, 2017]{Wolfinger2017SAS2017SE}
Wolfinger, R.~D. and Tan, P.-Y. (2017).
\newblock Sas-2017 stacked ensemble models for improved prediction accuracy.

\bibitem[Wolpert, 1992]{Wolpert1992}
Wolpert, D.~H. (1992).
\newblock Stacked generalization.
\newblock {\em Neural Networks}, 5(2):241--259.

\bibitem[Wright and Ziegler, 2017]{ranger}
Wright, M.~N. and Ziegler, A. (2017).
\newblock {ranger}: A fast implementation of random forests for high
  dimensional data in {C++} and {R}.
\newblock {\em Journal of Statistical Software}, 77(1):1--17.

\bibitem[Wundervald et~al., 2020]{Wundervald2020GeneralizingGP}
Wundervald, B.~D., Parnell, A.~C., and Domijan, K. (2020).
\newblock Generalizing gain penalization for feature selection in tree-based
  models.
\newblock {\em IEEE Access}, 8:190231--190239.

\bibitem[Zou and Hastie, 2005]{Zou2005}
Zou, H. and Hastie, T. (2005).
\newblock Regularization and variable selection via the elastic net.
\newblock {\em Journal of the Royal Statistical Society. Series B: Statistical
  Methodology}, 67:301--320.

\end{thebibliography}

\end{document}